\documentclass[final]{l4dc2026}

\title[Fourier Weak SINDy]{Fourier Weak SINDy: Spectral Test Function Selection for Robust Model Identification}
\usepackage{times}
\usepackage{bm}
\usepackage{hyperref}

\usepackage{xcolor}

\author{%
 \Name{Zhiheng Chen} \Email{zc548@cornell.edu}\\
 \addr Sibley School of Mechanical and Aerospace Engineering, Cornell University, Ithaca, New York, USA, 14850
 \AND
 \Name{Urban Fasel} \Email{u.fasel@imperial.ac.uk}\\
 \addr Department of Aeronautics, Imperial College London, London, United Kingdom
 \AND 
 \Name{Anastasia Bizyaeva} \Email{anastasiab@cornell.edu}\\
 \addr Sibley School of Mechanical and Aerospace Engineering, Cornell University, Ithaca, New York, USA, 14850
}

\begin{document}

\maketitle

\begin{abstract}%
 We introduce Fourier Weak SINDy, a minimal noise-robust and interpretable derivative-free equation learning method that combines weak-form sparse equation learning with spectral density estimation for data-driven test function selection. By using orthogonal sinusoidal test functions inspired by their prevalence in Modulating Function-based system identification, the weak-form sparse regression problem reduces to a regression over Fourier coefficients. Dominant frequencies are then selected via multitaper estimation of the frequency spectrum of the data. This formulation unifies weak-form learning and spectral estimation within a compact and flexible 
 framework. We illustrate the effectiveness of this approach in numerical experiments across multiple chaotic and hyperchaotic ODE benchmarks. 
\end{abstract}

\begin{keywords}%
  equation learning, nonlinear system identification, sparse regression, nonlinear dynamical systems, spectral density estimation%
\end{keywords}

\section{Introduction}

In this paper, we present a minimal interpretable framework for noise-robust derivative-free equation learning that combines recent advances in sparse system identification with classical signal processing tools to address the challenge of data-driven test function selection in the algorithm design. In recent years, sparse regression based methods for data-driven modeling have received significant attention. Among these, the Sparse Identification of Nonlinear Dynamics (SINDy) algorithm is a popular data-driven framework 
for learning 
Ordinary Differential Equations (ODEs), 
that uses sparsity-promoting optimization to identify unknown model coefficients while avoiding overfitting \citep{brunton2016discovering}. SINDy has been 
applied to equation learning problems in different domains from fluid dynamics to disease modeling~\citep{fukami2021sparse, delabays2025hypergraph, horrocks2020algorithmic}, and has found many extensions including learning partial differential equations~\citep{rudy2017data}, simultaneously learning coordinates and differential equations~\citep{champion2019data,bakarji2023discovering}, and quantifying uncertainty in the learned model structure and coefficients~\cite{zhang2018robust,north2022bayesian,niven2024dynamical,fung2025rapid}. 

Measurement noise and limited data are key challenges for learning models from data, motivating extensions of the SINDy framework such as ensembling~\citep{fasel2022ensemble}, denoising signals \citep{kaheman2022automatic}, and weak-form reformulations~\citep{schaeffer2017sparse, reinbold2020using, messenger2021weak, messenger2024weak,stephany2024weak}. The weak-form or integral-form family of methods are particularly promising noise-robust variants of SINDy, as they bypass the need for estimation of derivative information from noise corrupted data, which makes the original SINDy framework fragile. Weak-form methods instead rely on integrating data against user-defined analytic smooth test functions and their derivatives. Success of these methods therefore relies on selecting and tuning a good test function basis given characteristics of the collected data, to which significant attention has been devoted in recent work. For example, \cite{messenger2021weak} propose adaptive location and radius selections for compactly-supported ``bump''-shaped test functions based on approximate error covariance, the width-at-half-max parameter, and dominant wave modes; \cite{stephany2024weak} propose an optimization framework for test function tuning for weak-form PDE learning; \cite{bortz2023direct} start with compactly supported test functions and orthogonalize them using SVD, with the orthogonalized test functions resembling local Fourier basis or wavelets; \cite{gurevich2019robust} provide a detailed heuristic discussion of test function selection based on the frequency content of the data.

In modern weak-form sparse equation learning formulations, ``bump''-shaped test functions are a popular design choice due to their compact support and interpretable parameters. 
However, this choice introduces significant degrees of freedom into the test function selection problem, as the support region boundaries as well as the shape parameters of the test functions become optimization parameters. 
Furthermore, such test functions are not orthogonal, which means redundant information in the signal is reused in different components of the regression problem. 
In the work presented in this paper, we will illustrate that a simple test function basis of orthogonal sinusoids is an interpretable and robust choice for weak-form learning.
In this formulation, we take inspiration from classical Modulating Function Method system identification literature \citep{shinbrot1957analysis}, in which orthogonal test functions (or modulating functions) have been widely used for system identification \citep{pearson1985identification,pearson1990frequency,pearson1995aerodynamic}. The utility of orthogonal test functions has also recently been noted in the weak SINDy literature with data-driven constructions \citep{bortz2023direct} and derivation of general error bounds \citep{russo2024convergence}. 
Furthermore, we will show that with sinusoidal test functions, classical spectral density estimation (SDE) presents a natural strategy for data-driven selections of test function frequencies. 
Common approaches to SDE include nonparametric methods such as Bartlett method and Welch method \citep{stoica2005spectral}; a more advanced approach with higher resolutions and lower variances is the multitaper method \citep{thomson1982spectrum}. Due to its strong frequency specificity, the multitaper method in particular is popular across disciplines for the spectral analysis of complex time series data, from seismic data analysis to brain wave analysis \citep{park1987multitaper,van2007comparison,babadi2014review}. We will utilize multitaper to inform the selection of a sinusoidal test function basis.

\begin{figure}[!htb]
    \centering
    \includegraphics[width=1\linewidth,trim=0 0 18 0,clip]{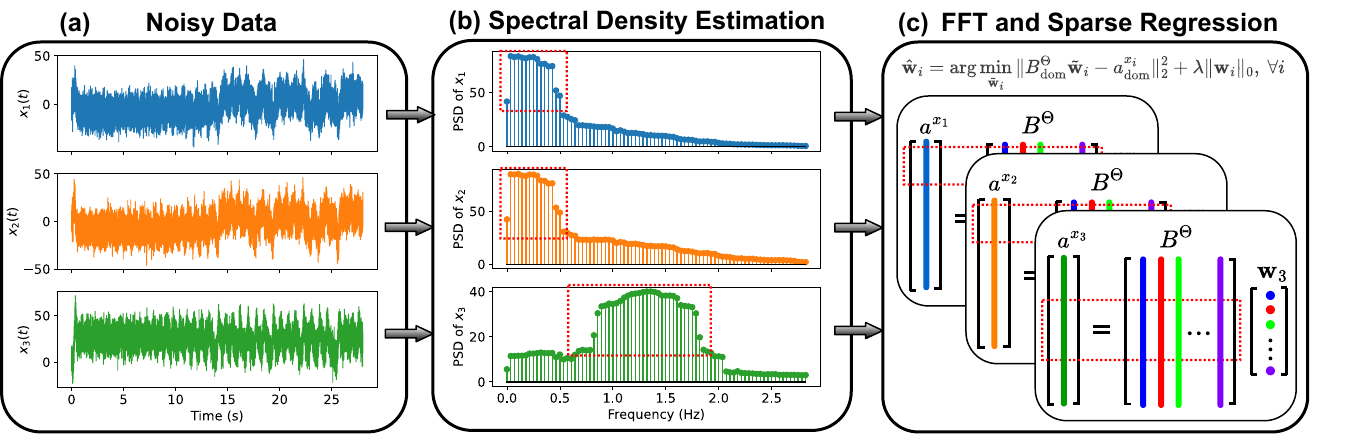}
    \caption{Graphical summary of proposed Fourier Weak SINDy method for sparse equation learning described in detail in Section \ref{sec:method}. First, multitaper spectral density estimation is used to identify dominant temporal frequencies in the observed noisy signals. Second, spectral sparse regression problem is solved to learn model coefficients in a selected function dictionary, leveraging the identified dominant Fourier frequencies as test functions in a weak-form system identification problem.}
    \label{fig: schematic}
\end{figure}

In this paper, we present the following contributions. First, we introduce Fourier Weak SINDy, a minimal method that combines weak-form sparse regression with spectral density estimation to learn parsimonious sparse ODE models from noisy measurement data. A high-level graphical summary of the method can be found in Figure \ref{fig: schematic}. We show that under the choice of sinusoidal test functions, the Weak SINDy sparse regression problem simplifies to a regression over Fourier series coefficients, which are efficiently computable with the Fast Fourier Transform (FFT). The choice of coefficients to regress over is made in a data-driven way by selecting dominant frequency modes in the data using multitaper estimation. Second, we illustrate effectiveness of this minimal approach in simulation studies. In particular, we provide numerical evidence that our method outperforms baseline SINDy and Weak SINDy methods across several chaotic and hyperchaotic ODE benchmarks for varying levels of measurement noise. Importantly, the SINDy and Weak SINDy baselines we compare against follow the implementations in the standard PySINDy package \citep{Kaptanoglu2022}. We leave principled comparisons against Weak SINDy variants with adaptive hyperparameter tuning, for example as in \cite{messenger2021weak}, to future work. Finally, we investigate the sensitivity of the method to the size of test function dictionary and the bandwidth of the spectral density estimation, illustrating that a reasonably small test function dictionary and bandwidth window yields robust performance on the low-dimensional ODE benchmarks. Our simulation code is available in a shared \href{https://github.com/Zhiheng-Chen/FourierWeakSINDy/}{Github repository}.

This paper is structured as follows. Section \ref{sec:background} reviews the fundamental formulations and limitations of current SINDy and weak SINDy frameworks; Section \ref{sec:method} presents our Fourier weak SINDy method, with mathematical derivations for the frequency-domain representation and the frequency selections based on spectral density estimations; Section \ref{sec:results} shows numerical experiment results, where Fourier weak SINDy is compared with standard SINDy and weak SINDy on equation learning tasks of four benchmark systems; Section \ref{sec:discussion} discusses limitations and future work. 


\section{Background \label{sec:background}}

Consider the problem of learning the structure of the governing equations for a finite-dimensional autonomous dynamical system given noisy observations of its state, 
\begin{align}
    \dot{\mathbf{x}} = \mathbf{f}(\mathbf{x}), \quad \quad \mathbf{y}(t) = \mathbf{x}(t) + \boldsymbol{\xi}(t) \label{eq:ODE}
\end{align}
where $\mathbf{x}(t) \in \mathbb{R}^n$, $\dot{\mathbf{x}}$ is the time derivative, $\mathbf{f}: \mathbb{R}^n \rightarrow \mathbb{R}^n$, and $\boldsymbol{\xi}(t) \in \mathbb{R}^n$ is a vector of additive measurement noise with components $\xi_i(t) \sim \mathcal{N}(0,\sigma^2)$ independent across state dimensions and time. We assume the output is sampled at regular time intervals $ t \in \{t_1, \dots, t_k\}$, $t_{i+1} - t_i = \Delta t$,  with the collected data snapshots summarized in the matrix
\begin{equation}
    Y = \begin{bmatrix} \mathbf{y}(t_1) & \dots & \mathbf{y}(t_k) \end{bmatrix}^T \in \mathbb{R}^{k \times n}. \label{eq:data-matrix}
\end{equation}

\subsection{Sparse Identification of Nonlinear Dynamics (SINDy) \label{sec:SINDy}}

The Sparse Identification of Nonlinear Dynamics (SINDy) method for equation learning has received significant interdisciplinary attention \citep{brunton2016discovering,brunton2025machine}. SINDy combines classical least-squares system identification \citep{aastrom1971system,unbehauen1990continuous} with a sparsity-promoting sequentially thresholded least squares solver to identify the coefficients of nonlinear terms in the function $\mathbf{f}(\mathbf{x})$ in \eqref{eq:ODE} from a large dictionary of candidate functions, under the assumption that $\mathbf{f}(\mathbf{x})$ is linear-in-coefficients and sparse in the selected dictionary.

To set up the sparse identification problem, define $m$ dictionary functions $\{\theta_1, \dots, \theta_m\}$ where each $\theta_i: \mathbb{R}^n \to \mathbb{R}$. Common dictionary choices include polynomials up to degree $d$, i.e. $\{\theta_j(\mathbf{x})\}_{j=1}^m = \{1, x_1, \dots, x_n, x_1^2,$ $ x_1x_2, \dots, x_n^d\}$, or other function bases such as trigonometric functions. It is assumed that each component $f_i(\mathbf{x})$ of the vector field $\mathbf{f}(\mathbf{x}) = [f_1(\mathbf{x}), \dots, f_n(\mathbf{x})]^T$ admits a representation
\begin{align}
    f_i(\mathbf{x}(t)) = \sum_{j=1}^m w_{ij} \theta_j(\mathbf{x}(t)) = \Theta(\mathbf{x}(t)) \mathbf{w}_i 
    \label{eq: library expansion}
\end{align} where $\mathbf{w}_i \in \mathbb{R}^m$ is a coefficient vector and $\Theta(\mathbf{x}(t)) = \begin{bmatrix} \theta_1(\mathbf{x}(t)) &  \dots &  \theta_m(\mathbf{x}(t)) \end{bmatrix} \in \mathbb{R}^{1 \times m}$. The true coefficient matrix $W = \begin{bmatrix} \mathbf{w}_1 & \dots & \mathbf{w}_n\end{bmatrix} \in \mathbb{R}^{m \times n}$ parametrizes the right hand side of \eqref{eq:ODE} with respect to the function dictionary as $\mathbf{f}(\mathbf{x}(t)) = \Theta(\mathbf{x}(t)) W$. 
The objective is to recover the unknown coefficients $W$ from the observed snapshots \eqref{eq:data-matrix} under the assumption that each $\mathbf{w}_i$ is sparse, i.e. each function component $f_i(\mathbf{x}(t))$ comprises only a small number of dictionary terms.

Given time series $Y$ of state measurements defined in \eqref{eq:data-matrix}, an estimate of the corresponding time derivatives of the state $\dot{X} = \begin{bmatrix}\dot{\mathbf{x}}(t_1) & \dots & \dot{\mathbf{x}}(t_k) \end{bmatrix}^T \in \mathbb{R}^{k \times n}$ is constructed, for example using finite-difference approximations or polynomial smoothing.  By stacking evaluations of the dictionary functions at each snapshot, 
we can construct the dictionary matrix 
\begin{equation} \Theta(Y) = \begin{bmatrix} \Theta(\mathbf{y}(t_1) \\ \vdots \\ \Theta(\mathbf{y}(t_k)) \end{bmatrix} = \begin{bmatrix} \theta_1(\mathbf{y}(t_1)) & \dots & \theta_m(\mathbf{y}(t_1) \\
\vdots & \ddots & \vdots \\
\theta_1(\mathbf{y}(t_k)) & \dots & \theta_m(\mathbf{y}(t_k))\end{bmatrix} \in \mathbb{R}^{k \times m}. \label{eq:dictionary_matrix}
\end{equation}
Approximation of the unknown coefficient matrix $W$ 
can be formulated as the least squares problem
\begin{align}
    \widehat{W} = \arg \min_{\tilde{W}} \left\| \dot{X}-\Theta(Y) \tilde{W} \right\|_2^2.
    \label{eq: SINDy LS}
\end{align}
The regression problem  \eqref{eq: SINDy LS} is then solved using a sparsity promoting method such as sequentially thresholded least squares or sequentially thresholded ridge regression as proposed in \citep{brunton2016discovering,rudy2017data}; that is, we iteratively solve \eqref{eq: SINDy LS} using standard least squares or ridge regression, and at each iteration setting to zero coefficients whose magnitudes fall below a specified threshold $\eta > 0$. After $K$ iterations, the resulting coefficient matrix $\widehat{W}$ identifies active dictionary terms in each component of the vector field \eqref{eq: library expansion}.

\subsection{Weak SINDy and Test Function Selection \label{sec:WSINDy}}
While SINDy is an efficient and interpretable framework for model discovery, its reliance on derivative estimates $\Dot{X}$ from noisy measurements limits its robustness. In practice, SINDy coefficient estimates become highly inaccurate even with reasonably large signal-to-noise ratios in the data matrix \eqref{eq:data-matrix}. 
To mitigate this noise sensitivity, a number of recent research efforts proposed integral-form or weak-form reformulations of SINDy, which bypass the need for estimating derivatives \citep{schaeffer2017sparse,gurevich2019robust,reinbold2020using,messenger2021weak,bortz2023direct,stephany2024weak}. 
In the Weak SINDy formulation, each component $f_i(\mathbf{x})$ of \eqref{eq:ODE} is projected onto a compactly-supported test function $\phi_j \in C^1_c([t_a,t_b])$ and integrated over its support region, $\int_{t_a}^{t_b} \dot{x}_i(t) \phi_j(t) dt = \int_{t_a}^{t_b} f_i(\mathbf{x}(t))\phi_j(t) dt$.
Imposing that test functions vanish at the end points of the time interval, i.e. $\phi_j(t_a) = \phi_j(t_b) = 0$, and applying integration-by-parts yields
\begin{align}
    -\int_{t_a}^{t_b} x_i(t) {\phi}_j'(t) dt
    =
    \int_{t_a}^{t_b} f_i(\mathbf{x}(t))\phi_j(t) dt.
    \label{eq: weak derivative}
\end{align}
In discrete time, the integrals in \eqref{eq: weak derivative} are approximated using numerical quadrature. The matrices of evaluations of $\phi_j(t)$ and its derivative $\phi_j'(t)$ at the sampled time points are defined as $\Phi_j = \begin{bmatrix} \phi_j(t_1) & \dots \phi_j(t_k) \end{bmatrix} $ $ \in \mathbb{R}^{1 \times k}$ and $\Phi_j'= \begin{bmatrix} \phi_j'(t_1) & \dots \phi_j'(t_k) \end{bmatrix}\in \mathbb{R}^{1 \times k}$. Defining $X \in \mathbb{R}^{k \times n}$ as the data snapshot matrix and leveraging the library matrix expansion \eqref{eq: library expansion} along with the dictionary matrix definition \eqref{eq:dictionary_matrix}, the integral expression \eqref{eq: weak derivative} becomes
$ -\Delta t \Phi'_j X = \Delta t \Phi_j \Theta(X) W$.
Collecting evaluations of $p$ distinct test functions into the matrices $\Phi \in \mathbb{R}^{p \times k}$ and $\Phi' \in \mathbb{R}^{p \times k}$, and incorporating noisy data observations \eqref{eq:data-matrix}, coefficients $W$ of the ODE are then estimated by solving 
\begin{align}
    \widehat{W} = \arg\min_{\tilde{W}}
              \left\| \Delta t \Phi \Theta(Y) \tilde{W} + \Delta t \Phi' Y \right\|_2^2. \label{eq: LS_WSINDy}
\end{align}
As in the standard SINDy formulation, coefficient sparsity is enforced by solving \eqref{eq: LS_WSINDy} using least squares or ridge regression with sequential thresholding.

A key design choice that significantly affects the performance of weak-form methods is the form of test function $\phi_j(t)$. Grounded in the definition of a weak derivative, test functions $\phi_j$ 
are often constrained to be compactly supported over the observation interval. Typical test functions have the form of a localized ``bump'', for example \cite{reinbold2020using} utilize functions of the form
\begin{align}
    \phi_j(t) = (1-\underline{t}^2)^q
    \label{eq: bump}
\end{align}
where $q \in \mathbb{Z}^+$ is a hyperparameter (larger $q$ gives narrower bumps around the subdomain center), and $\underline{t}$ is the dimensionless time obtained via $\underline{t} = (t-t_j)/H_t$ with $t_j$ being the center of the $j$-th subdomain and $H_t$ being the duration of each subdomain. Optimizing the performance of weak SINDy then requires strategic selections of the sharpness and support interval boundaries of the test functions. For example, \cite{messenger2021weak,bortz2023direct} present an adaptive-grid strategy that places more test functions at locations where large changes in the dynamics occur, where the gradient is approximated by formulating and solving a nonlinear root finding problem; then they select hyperparameters and support intervals based on the approximate error covariance, the width-at-half-max parameter, and optionally the dominant wave modes.

\section{Fourier Weak SINDy \label{sec:method}}
The utility of orthogonal test functions has been explored in several prior works. For example, \cite{bortz2023direct} construct test functions by orthogonalizing smooth bump functions via Singular Value Decomposition (SVD). Moreover, in classical systems and control literature, such as \cite{pearson1985identification,pearson1990frequency,pearson1995aerodynamic}, Fourier-mode test functions (also known as modulating functions) are used for identification of linear systems, without requiring them to be compactly supported, and only requiring that they should vanish at domain boundaries.
Motivated by these precedents, in this section we introduce sinusoidal test functions into the weak SINDy framework and illustrate that the resulting learning problem reduces to a sparse regression over Fourier coefficients.
This reformulation also reinterprets test function selection as the selection of imformative Fourier frequencies, which can be addressed using classical signal processing methods for spectral density estimation. Together, this formulation yields a noise-robust and interpretable spectral approach to sparse equation learning.


\subsection{Sinusoidal Test Functions}

Suppose the data we collect has duration $T$ (i.e., $T = t_k-t_1$), then define the $\ell$-th sinusoidal test function and its derivative, where $\ell \in \mathbb{Z}^+$, as 
\begin{equation}
         \phi_\ell(t) = \sin\left(\frac{2\pi \ell}{T}t\right), \quad \quad
        \phi'_\ell(t) = \frac{2\pi \ell}{T}\cos\left(\frac{2\pi \ell}{T}t\right).
        \label{eq: sine test func}
\end{equation}
Since $\phi_\ell(0) = \phi_\ell(T) = 0$ for all $\ell \in \mathbb{Z}^+$, the weak form equation \eqref{eq: weak derivative} becomes 
\begin{align}
    -\frac{2\pi \ell}{T}\int_0^T x_i(t) \cos\left(\frac{2\pi \ell}{T}t\right) dt
    = 
    \int_0^T f_i(\mathbf{x}(t)) \sin\left(\frac{2\pi \ell}{T}t\right) dt.
    \label{eq: Fourier weak form}
\end{align}

Let $x_i(t)$ and $f_i(t)$ defined over $[0,T]$ be expanded as Fourier series 
\begin{subequations}
    \begin{align}
            x_i(t) = a_0^{x_i}+ \sum_{\ell = 1}^{\infty} \left( a_\ell^{x_i} 
                     \cos\left(\frac{2\pi \ell}{T}t\right) + b_\ell^{x_i} 
                     \sin\left(\frac{2\pi \ell}{T}t\right) \right), \\
            f_i(x(t)) = a_0^{f_i}  + \sum_{\ell = 1}^{\infty} \left( a_\ell^{f_i} 
                        \cos\left(\frac{2\pi \ell}{T}t\right)+ b_\ell^{f_i} 
                        \sin\left(\frac{2\pi \ell}{T}t\right)\right).
    \end{align}
    \label{eq: F-series of dynamics}
\end{subequations}
Since the Fourier basis are mutually orthogonal, substituting \eqref{eq: F-series of dynamics} into \eqref{eq: Fourier weak form} lets most of the inner products vanish, giving
\begin{align}
    -\frac{2\pi \ell}{T} a_\ell^{x_i} = b_\ell^{f_i}, \quad \text{for all} \ \ell \in
    \mathbb{Z}^+.
    \label{eq: dynamics in Fourier coeffs}
\end{align}
Note that the $a_\ell^{x_i}$ and $b_\ell^{f_i}$ coefficients can be efficiently computed using FFT of the data matrix $Y$ and the dictionary matrix $\Theta(Y)$. 

\subsection{Dictionary Function Expansion}
In the equation learning problem, we do not know the exact form of $f_i$ and instead expand it in a dictionary of functions $\Theta(\mathbf{x}) = \begin{bmatrix} \theta_1(\mathbf{x}) &  \dots &  \theta_m(\mathbf{x}) \end{bmatrix}$, with the relation described in \eqref{eq: library expansion}. Let the Fourier series expansion of $\theta_j(\mathbf{x}(t))$ defined over $[0,T]$ be
\begin{align}
    \theta_j(\mathbf{x}(t)) 
    = 
    a_0^{\theta_j}  + \sum_{\ell = 1}^{\infty} \left( a_\ell^{\theta_j} \cos\left(\frac{2\pi \ell}{T}t\right) + b_\ell ^{\theta_j} 
    \sin\left(\frac{2\pi \ell}{T}t\right) \right).
    \label{eq: F-series on theta_l}
\end{align}
By linearity of the Fourier operator, the right-hand side of \eqref{eq: dynamics in Fourier coeffs} can then be written as
\begin{align}
    b_\ell^{f_i} 
    = 
    \sum_{j = 1}^m w_{ij} b_\ell^{\theta_j}.
\end{align}
Define $B_\ell^\Theta = \begin{bmatrix} b_\ell^{\theta_1} & \ldots & b_\ell^{\theta_m} \end{bmatrix}$; then $b_\ell^{f_i} = B_\ell^\Theta \mathbf{w}_i$, and \eqref{eq: dynamics in Fourier coeffs} becomes
\begin{align}
    -\frac{2\pi \ell}{T} a_\ell^{x_i} = B_\ell^\Theta \mathbf{w}_i.
\end{align}
Since for each $j$, $b_\ell^{\theta_j}$ for all $\ell$ up to the index for Nyquist frequency can be efficiently computed using one FFT, this formulation can be significantly faster to implement than quadrature-based approaches, where an evaluation of quadrature is needed for each test function.

\subsection{Frequency Selection Based on Spectral Density Estimation}
The multitaper method provides a low-variance, low-leakage, and high-resolution approach to spectral density estimation (SDE) compared to simple single-taper approaches~\cite{thomson1982spectrum}. 
Therefore, we use the multitaper method to perform SDE for trajectory data in order to select informative spectral test function frequencies. A brief overview of the multitaper method with Slepian sequences is available in the supplement.



For each $f_i$, we apply the multitaper method to estimate the power spectral density of $x_i(t)$ from noisy measurements, and select frequencies at which the estimated power is highest, i.e. the dominant frequencies. At dominant frequencies, the signal-to-noise ratio is highest, making the corresponding spectral content the most reliable basis for system identification from noisy data. 
Suppose we pick $K$ dominant frequencies in total, and define
\begin{align}
    a^{x_i}_\text{dom} 
    = 
    \begin{bmatrix} 
        -\frac{-2\pi}{T}a^{x_i}_1 & \ldots & -\frac{-2\pi K}{T}a^{x_i}_K 
    \end{bmatrix}^T, \quad 
    B^\Theta_\text{dom} = \begin{bmatrix}
                    (B_1^{\Theta})^T & \ldots  & (B_K^{\Theta})^T
               \end{bmatrix}^T.
\end{align}
Then we can obtain, and solve using sequentially thresholded methods, the least squares problem
\begin{align}
    \widehat{\mathbf{w}}_i
    =
    \arg \min_{\tilde{\mathbf{w}}_i}
    \left\| B^\Theta_\text{dom} \tilde{\mathbf{w}}_i - a^{x_i}_\text{dom} \right\|_2^2.
\end{align}
This approach to test function selection is interpretable in a physically meaningful sense, since the selected frequencies correspond directly to the dominant  modes of the observed signal. The choice of test functions is directly grounded in the spectral structure of the data, rather than indirectly through abstract optimization criteria, and utilizes classical signal processing methods.


%

\section{Simulation Results \label{sec:results}}



We implement the Fourier Weak SINDy framework in Python, testing it on simulated trajectories of several benchmark nonlinear dynamical systems with varying measurement noise levels. The trajectory data have durations of 10 s and are evenly sampled with 1000 samples per second. We follow the noise level definition and error metrics in \cite{messenger2021weak}. Specifically, the noise ratio $\sigma_{NR}$ is defined using
\begin{align}
    \sigma = \sigma_{NR}\frac{\|X\|_F}{\sqrt{kn}}
\end{align}
where $\|X\|_F$ is the Frobenius norm of the trajectory data matrix $X \in \mathbb{R}^{k \times n}$, and the noise level is then $\sigma_{NR} \times 100 \%$. We use two metrics to evaluate the error of equation learning results; the first metric is the relative coefficient error $E_2(\hat{W})$, defined as
\begin{align}
    E_2(\hat{W}) = \frac{\|\hat{W}-W\|_F}{\|W\|_F}.
\end{align}
The second metric is the true positive ratio (TPR)
\begin{align}
    \text{TPR}(\hat{W}) = \frac{\text{TP}(\hat{W})}
                          {\text{TP}(\hat{W})+\text{FP}(\hat{W})
                          +\text{FN}(\hat{W})}
\end{align}
where $\text{TP}(\hat{W})$ is the number of correctly identified nonzero terms, $\text{FP}(\hat{W})$ is the number of falsely identifd nonzero terms, and $\text{FN}(\hat{W})$ is the number of terms that are falsely identified as having a coefficient of zero. 

In the equation learning of all benchmark systems using Fourier weak SINDy, we use the sequentially thresholded ridge regression with a sparsity threshold of 0.5 and a ridge regularization value of 0.001. We implement two frequency selection methods for Fourier weak SINDy -- the first method is to use the first 500 frequencies of the FFT result (i.e., using a sweep of frequencies up to 50 Hz, $\frac{1}{10}$ of the Nyquist frequency), and the second method is to use the top 100 dominant frequencies from the SDE. We compare the results of Fourier weak SINDy with those of standard SINDy and weak SINDy methods in the PySINDy package \cite{Kaptanoglu2022}, where we choose sparse relaxed regularized regression to obtain sparse coefficient matrices, and use 1000 subdomains for the weak SINDy in PySINDy (i.e., 1000 test functions of the form \eqref{eq: bump}).

\begin{figure}[!htb]
    \includegraphics[width=1\linewidth,trim=2 10 8 6,clip]{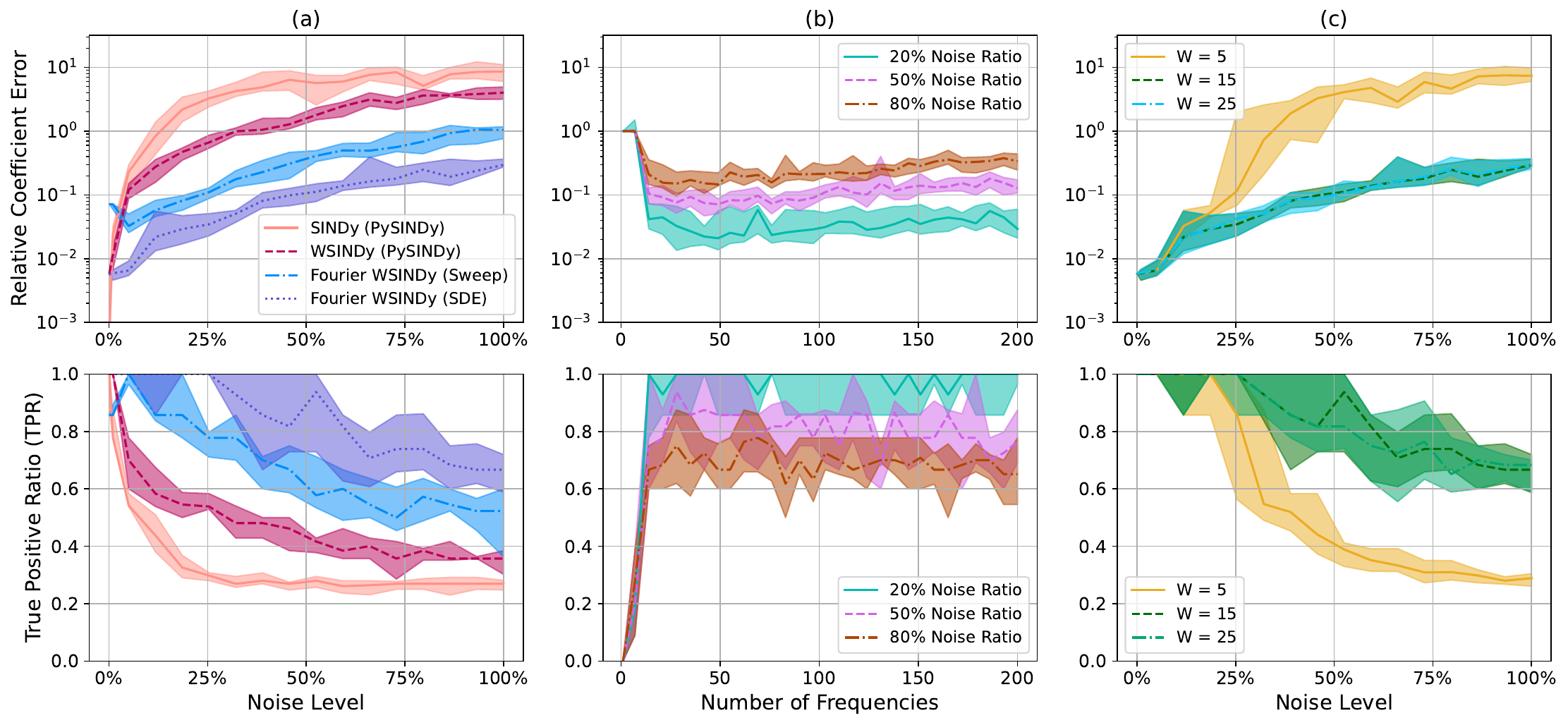}
    \caption{Lorenz system equation learning results with second-order polynomial library. Median and quartiles of the relative coefficient error and TPR of (a) four different SINDy and weak SINDy methods at varying noise levels, (b) Fourier weak SINDy with SDE and different numbers of dominant frequencies chosen, and (c) Fourier weak SINDy with different bandwidth settings (in Hz) for the multitaper SDE.}
    \label{fig: Lorenz2}
\end{figure}

We perform our first numerical experiment on the Lorenz system with library terms up to second-order polynomials. We set the initial conditions of the simulation to $\mathbf{x}_0 = \begin{bmatrix} 20 & 12 & -30 \end{bmatrix}^T$. As shown in Fig.~\ref{fig: Lorenz2}, we first compare the performances of SINDy (PySINDy), weak SINDy (PySINDy), and Fourier weak SINDy based on both the frequency sweep and the spectral density estimation methods,  with noise ratios ranging from 0.0001\% to 100\% and 20 noise instances per noise level. Moreover, we investigate the effect of choosing different numbers of dominant frequencies for the Fourier weak SINDy at different noise levels (the dominant frequencies are extracted using the FFT on the clean data, to eliminate the effect of SDE error and focus on the effect of frequency numbers), as well as how the bandwidth of the multitaper SDE influences equation learning results.

As shown in Fig.~\ref{fig: Lorenz2}(a) for the Lorenz system benchmark, at 100\% noise level, SINDy had a median coefficient error at the order of 10 and a median TPR near 0.2. This illustrates that the classical SINDy approach is not providing meaningful equation learning results at high measurement noise. In contrast, the Fourier weak SINDy approach with SDE achieved a median coefficient error on the order of 0.1 and a median TPR near 0.7. This implies that even with significant measurement noise, Fourier weak SINDy can still reveal the core structure of the model with reasonably small coefficient error. Furthermore, up through a 25\% noise level, Fourier Weak SINDy had a TPR of 1 across all of the numerical experiments, meaning that the sparse regression problem yielded no false positives or false negatives. This is significant improvement over the SINDy and Weak SINDy baselines, and over the Fourier Weak SINDy method with a sweep frequency selection. In addition to coefficient error and TPR, we evaluated simulated trajectory error for the Lorenz system, illustrated in a figure  available in the Supplement. We find that Fourier Weak SINDy outperforms the baseline methods, and is less likely to learn unstable dynamics.

Fig.~\ref{fig: Lorenz2}(b) illustrates the performance Fourier weak SINDy against the number of test functions being selected. Once the number of test functions is more than 30, the coefficient error and true positive ratio does not vary significantly as the number of test functions increases, though there is a slight decline in performance as test functions of higher frequencies are included. The slight decline in performance is likely caused by the fact that the frequency content of the signal is mostly in the top 50 to 100 dominant frequencies, and thus the components projected onto the less-dominant test functions contain more information from measurement noise that is not dynamically meaningful. Fig.~\ref{fig: Lorenz2}(c) reveals the performance of Fourier weak SINDy against the selection of the bandwidth of the multitaper SDE. The performance improves with the size of the bandwidth at first, but the benefit saturates as the bandwidth increases. 

\begin{figure}[!htb]
    \centering
    \includegraphics[width=1\linewidth,trim=2 10 6 6,clip]{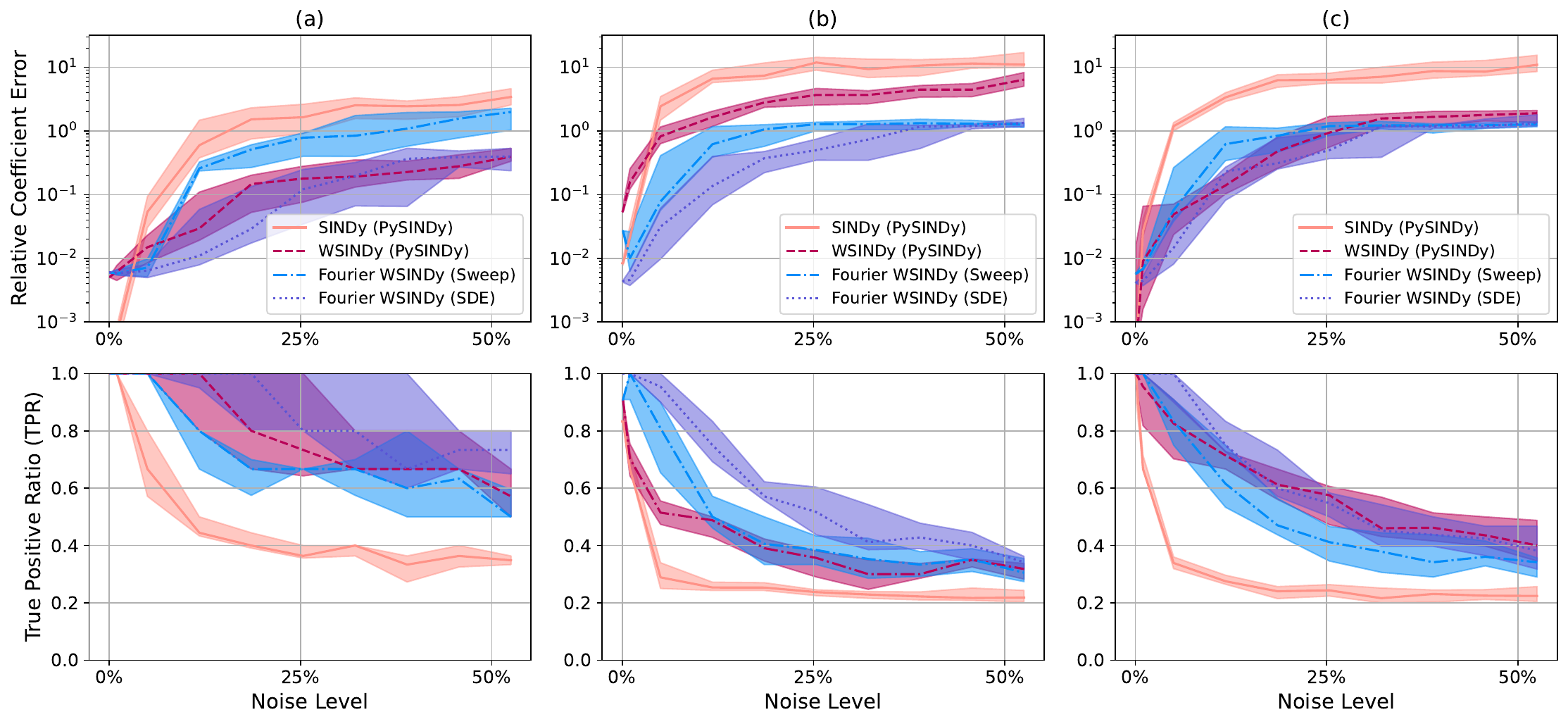}
    \caption{Equation learning results of four different SINDy and weak SINDy methods for (a) Lotka-Volterra equations, (b) hyperchaotic Lorenz system, and (c) hyperchaotic Jha system.}
    \label{fig: other2}
\end{figure}

We also perform numerical experiments on the Lotka-Volterra equations, the hyperchaotic Lorenz system, and the hyperchaotic Jha system, all with second-order polynomial libraries; the equation learning results are shown in Fig.~\ref{fig: other2}. The governing equations of all four benchmark systems, parameters and initial conditions used for benchmarking, and equation learning results with higher-order polynomial libraries are available in the supplement. 
As shown in Fig.~\ref{fig: other2}, Fourier weak SINDy with SDE typically improves on, or at worst performs comparably to, the baseline methods on these systems according to the equation error  and TPR metrics. 


\section{Discussion \label{sec:discussion}}
In this paper, we showed that combining weak-form equation learning over orthogonal sinusoidal test functions with classical signal processing techniques for the selection of test function frequencies to match maximally energetic modes in the data, presents an interpretable, robust, and effective approach to equation learning. We presented numerical evidence that Fourier weak SINDy shows either improved or comparable performance in equation learning tasks across four chaotic and hyperchaotic benchmark systems, compared to baseline methods including the ``bump" weak SINDy method in the PySINDy package. Moreover, the additional hyperparameters introduced by the method over the baselines, i.e. the number of dominant frequencies to be included in the regression problem and the bandwidth selection for the multitaper SDE, are interpretable and were shown numerically to have a saturating effect on method performance, providing practitioners with a simple and intuitive strategy for tuning. 

There are many promising directions of future work. For example, the current FFT implementations of Fourier weak SINDy are limited to uniformly sampled trajectory data. Therefore, a future direction is to experiment with Nonuniform Fast Fourier Transform to perform equation learning using data with irregular time samples. Moreover, an assumption made in this paper is that the measurement noise is additive white Gaussian noise. Although this assumption holds for many equation learning tasks, noise profiles beyond white noise are common in practice. In these cases, Fourier weak SINDy can be further augmented by more advanced spectral density estimation methods, such as the adaptive multitaper method and the Capon estimator. Additionally, existing techniques for improving the accuracy and noise-robustness of SINDy, such as ensembling and Bayesian coefficient selection strategies, can be incorporated into the current framework to further improve accuracy and robustness. Finally, the current framework utilizes spectral density estimation for test function selection; while the multitaper method effectively captures the dominant frequencies in terms of energy, dynamically meaningful information may be contained in low-amplitude modes of the signal. In this case, alternative test function selection strategies may be more effective, for example 
selection based on the Fourier Phase Index metric proposed by \cite{aguilar2024fourier}.






\acks{The authors are grateful to Daniel Messenger for helpful literature pointers to recent work on weak SINDy with orthogonal test functions.}

\bibliography{references}

\end{document}


\maketitle 

\section{Multitaper Spectral Density Estimation}
In this section, we review classical ideas from spectral analysis of signals \citep{stoica2005spectral}, which are used in the spectral density estimation component of our method. Given a right-sided random signal sequence $y(n)$ with length $N$, its power spectral density (PSD) $S(\omega)$ is defined as 
\begin{align}
    S(\omega) 
    = 
    \lim_{N\rightarrow\infty}
    E\left\{\frac{1}{N}\left|\sum_{n=0}^{N-1} y(n)e^{-\im\omega n}\right|^2 \right\}
\end{align}
where $E$ denotes expectation and $\im$ is the imaginary unit. Note that $\sum_{n=1}^N y(n)e^{-\im\omega n}$ is the discrete-time Fourier transform of the right-sided $y(n)$; therefore, $S(\omega)$ is a measurement of the amount of energy in $y(n)$ at frequency $\omega$, and the function $S$ represents the distribution of $y(n)$'s energy spanning a continuous spectrum of frequencies, which is a direct result from Parseval's theorem.

Direct computation of the PSD is not possible since it requires an infinite number of ensembles for $y(n)$, and thus spectral density estimations (SDEs) are needed. A common approach to SDE is via the periodogram, defined as
\begin{align}
    \hat{S}_P(\omega) 
    = 
    \frac{1}{N}\left|\sum_{n=0}^{N-1} y(n)e^{-\im\omega n}\right|^2.
\end{align}

The periodogram by itself cannot provide decent spectral density estimations due to its high variance. To obtain a consistent PSD estimate, the multitaper method averages modified periodograms obtained using a set of orthogonal window functions (tapers); a typical set of tapers is the Discrete Prolate Spheroidal (Slepian) Sequences \citep{thomson1982spectrum,slepian1978prolate}. Denote the $j$-th Slepian sequence as $g_j$, and denote the bandwidth normalized by the sampling rate as $W$, then $g_j$ can be defined by its convolution with the sinc function:
\begin{align}
    \sum_{m=0}^{N-1} \frac{\sin \left( 2\pi W (n-m) \right)}{\pi (n-m)} g_j(m) 
    = 
    \lambda_k g_j(n), \quad n,j = 0,1,2,\ldots,N-1,
\end{align}
and $g_j$'s can be obtained by solving the eigenvalue problem for the sinc function matrix. Then the $j$-th modified periodogram is constructed by first tapering the signal with $g_j$, then computing the squared magnitude of its discrete-time Fourier transform:
\begin{align}
    \hat{S}_j(\omega) 
    = 
    \frac{1}{N}\left|\sum_{n=0}^{N-1} g_j(n) y(n)e^{-\im\omega n}\right|^2
    \label{eq: S_hat_j}
\end{align}
There are approximately $2NW$ sequences with eigenvalues close to unity, and thus we choose $M = 2NW$ tapers. Then the PSD estimate from the multitaper method is the average of the $M$ modified periodograms:
\begin{align}
    \hat{S}_{MT}(\omega)
    =
    \frac{1}{M}\sum_{j=0}^{M-1} \hat{S}_j(\omega)
\end{align}

Note that the sinc function is the impulse response of an ideal lowpass filter with passband $[-W,W]$, and thus choosing a larger bandwidth will result in the tapers having a wider frequency content, causing the tapers to have a stronger frequency-averaging effect that can be visualized if we convert the Fourier transform of the time-domain multiplication in \eqref{eq: S_hat_j} into frequency-domain convolution. Therefore, if we choose a smaller and more concentrated bandwidth, then the PSD estimate will have more spectral leakage but higher resolution; conversely, if we choose a larger and more distributed bandwidth, then the PSD estimate will have less leakage but lower resolution.

\section{Benchmark Systems} 
In this section we define the structure, parameters, and initial conditions for ODE systems used to evaluate our equation learning method, with equations for the hyperchaotic benchmark systems chosen from \cite{gilpin2023chaos}. The Lorenz system has governing differential equations
\begin{subequations}
    \begin{align}
        & \dot{x} = \sigma(y-x) \\
        & \dot{y} = x(\rho-z)-y \\
        & \dot{z} = xy-\beta z
    \end{align}
\end{subequations}
where we use $\sigma = 10$, $\rho = 28$ $\beta = 8/3$, and initial condition $\begin{bmatrix} 20 & 12 & -30 \end{bmatrix}^T$ for benchmarking.

The Lotka-Volterra equations are
\begin{subequations}
    \begin{align}
        & \dot{x} = 3x-\beta xy \\
        & \dot{y} = -6y+\beta xy
    \end{align}
\end{subequations}
where we use $\beta = 1$ and initial conditions $\begin{bmatrix} 1 & 1 \end{bmatrix}^T$.

The hyperchaotic Lorenz system has governing differential equations
\begin{subequations}
    \begin{align}
        & \dot{x} = a(y-x)+w \\
        & \dot{y} = -xz+cx-y \\
        & \dot{z} = -bz+xy \\
        & \dot{w} = dw-xz
    \end{align}
\end{subequations}
where we use $a = 10$, $b = 2.667$, $c = 28$, $d = 1.1$, and initial conditions $\begin{bmatrix} 5 & 8 & 12 & 21 \end{bmatrix}^T$.

The hyperchaotic Jha system has governing differential equations
\begin{subequations}
    \begin{align}
        & \dot{x} = a(y-x)+w \\
        & \dot{y} = -xz+bx-y \\
        & \dot{z} = xy-cz \\
        & \dot{w} = -xz+dw
    \end{align}
\end{subequations}
where we use $a = 10$, $b = 28$, $c = 8/3$, $d = 1.3$, and initial conditions $\begin{bmatrix} 0.1 & 0.1 & 0.1 & 0.1 \end{bmatrix}^T$.


\section{Simulation Results for Higher Order Polynomial Dictionaries}

\begin{figure}[H]
    \centering
    \includegraphics[width=0.98\linewidth,trim=2 10 6 6,clip]{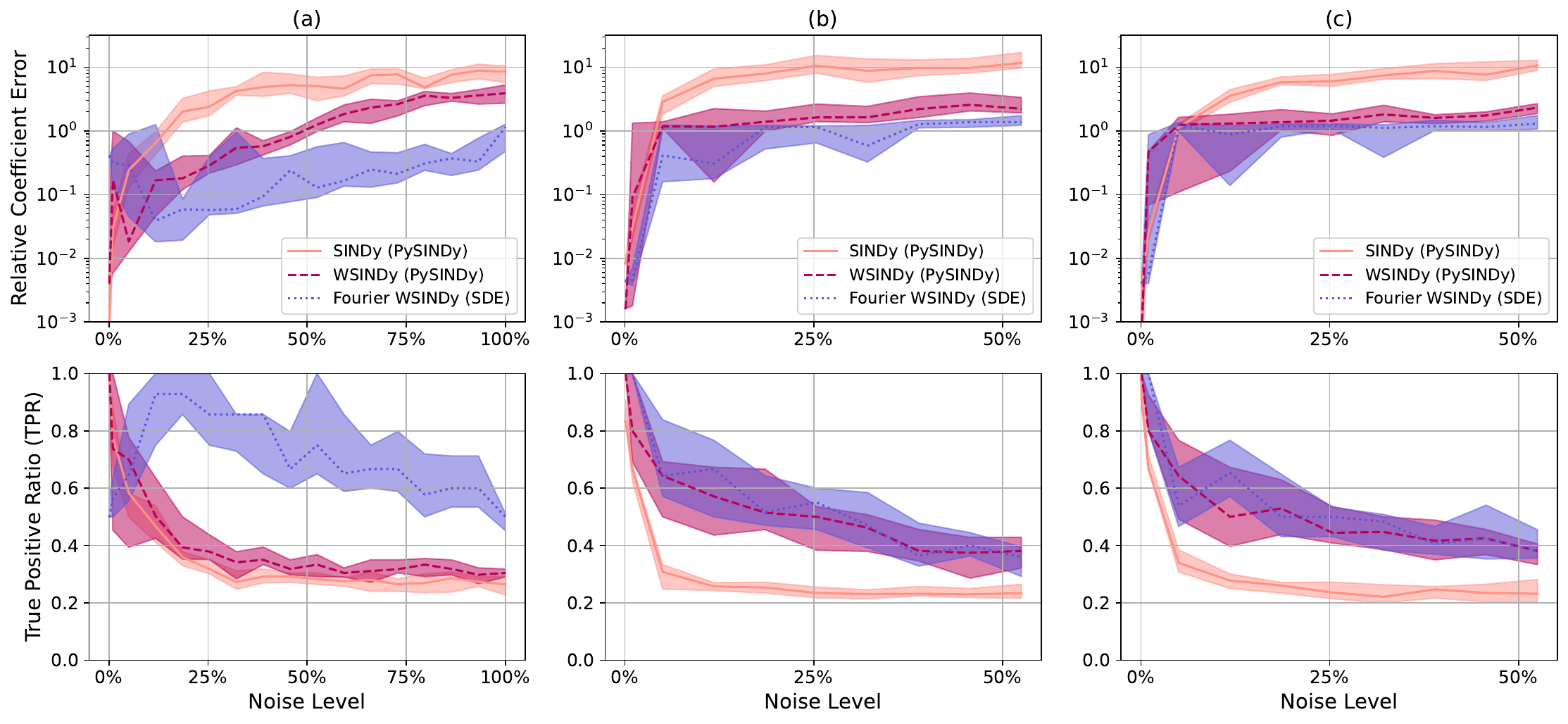}
    \caption{Equation learning results comparisons of (a) Lorenz system with fifth-order polynomial library, (b) hyperchaotic Lorenz system with third-order polynomial library, and (c) hyperchaotic Jha system with third-order polynomial library.}
    \label{fig: other2}
\end{figure}

In addition to numerical experiments presented in the main text, we conducted numerical experiments with higher-order polynomial dictionaries, shown in Fig. \ref{fig: other2}.

Note that in the experiments on the bump-function weak SINDy for the hyperchaotic Jha and hyperchaotic Lorenz systems with the third-order polynomial library, we choose sequentially-thresholded ridge regression (STRR) proposed by \cite{rudy2017data} rather than sparse relaxed regularized regression as STRR offers better accuracy in this experimental setting. 

\section{Simulation Results for Trajectory Error}
\begin{figure}[H]
    \centering
    \includegraphics[width=0.98\linewidth,trim=2 10 6 6,clip]{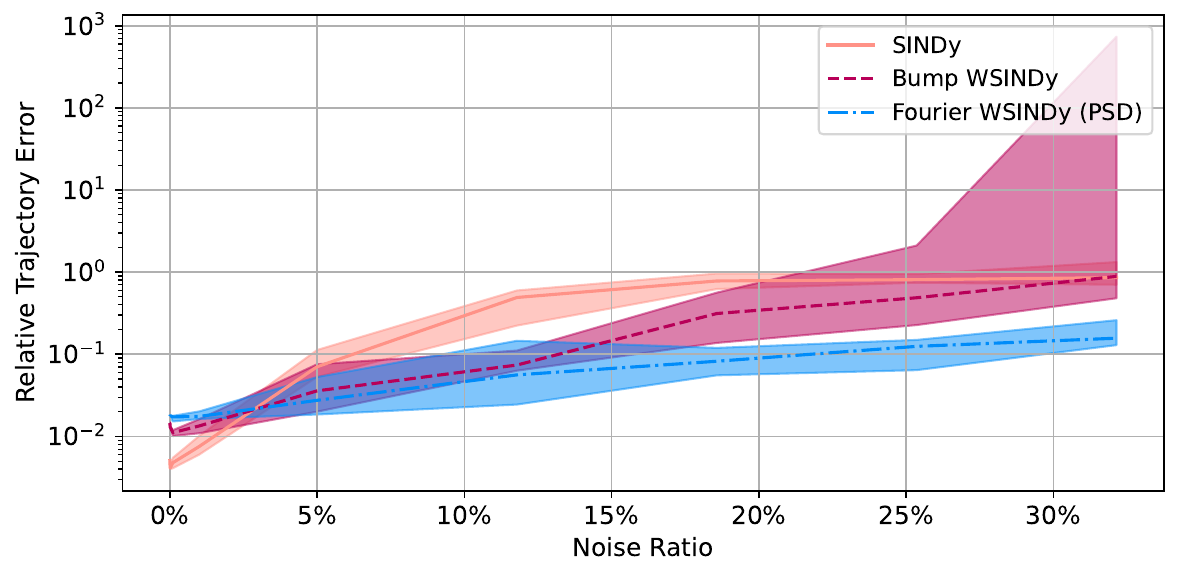}
    \caption{Trajectory error comparisons for regular SINDy, bump-function weak SINDy, and Fourier weak SINDy.}
    \label{fig: traj error}
\end{figure}

We conducted numerical simulations that compare the trajectory errors of regular SINDy, bump-function weak SINDy, and Fourier weak SINDy. The relative trajectory error is defined as 
\begin{align}
    \varepsilon_2 = \frac{\|\hat{X}-X_\text{true}\|_F}
                    {\|X_\text{true}\|_F}
\end{align}
where $\hat{X}$ is the simulated trajectory of a learned model and $X_\text{true}$ is the simulated trajectory of the true model. All tests are performed on the Lorenz system with second-order polynomial library. Out of 200 equation learning results from each SINDy variant (up to 30\% noise level), the regular SINDy and bump-function weak SINDy both learned 12 unstable dynamics models, while all dynamics models learned by Fourier weak SINDy are stable. The simulation results are shown in Fig.~\ref{fig: traj error}.

\bibliography{references}